\documentclass{article}

\usepackage{arxiv}
\usepackage[utf8]{inputenc} 
\usepackage[T1]{fontenc}    
\usepackage{hyperref}       
\usepackage{url}            
\usepackage{booktabs}       
\usepackage{amsfonts}       
\usepackage{nicefrac}       
\usepackage{microtype}      
\usepackage{mathtools}
\usepackage{graphicx}
\usepackage{amssymb}
\usepackage{physics}        
\usepackage[style=numeric-comp, sorting=none]{biblatex}

\title{MHT-X: Offline Multiple Hypothesis \\ Tracking with Algorithm X}

\author{
  Peteris Zvejnieks\\
  Institute of Numerical Modelling\\
  University of Latvia (UL)\\
  Riga, Latvia, Jelgavas 3, 1004 \\
  \texttt{peteris.zvejnieks@lu.lv} \\
   \And
  Mihails Birjukovs\\
  Institute of Numerical Modelling\\
  University of Latvia (UL)\\
  Riga, Latvia, Jelgavas 3, 1004 \\
  \texttt{mihails.birjukovs@lu.lv} \\
   \And
   Martins Klevs\\
  Institute of Numerical Modelling\\
  University of Latvia (UL)\\
  Riga, Latvia, Jelgavas 3, 1004 \\
   \And
   Megumi Akashi\\
  Helmholtz-Zentrum Dresden-Rossendorf (HZDR)\\
  Department of Magnetohydrodynamics\\
  Bautzner Landstraße 400, 01328 Dresden, Germany \\
   \And
    Sven Eckert\\
  Helmholtz-Zentrum Dresden-Rossendorf (HZDR)\\
  Department of Magnetohydrodynamics\\
  Bautzner Landstraße 400, 01328 Dresden, Germany \\
   \And
    Andris Jakovics\\
  Institute of Numerical Modelling\\
  University of Latvia (UL)\\
  Riga, Latvia, Jelgavas 3, 1004
  }

\begin{document}
	\maketitle

	\begin{abstract}
		An efficient and versatile implementation of offline multiple hypothesis tracking with Algorithm X for optimal association search was developed using Python. The code is intended for scientific applications that do not require online processing. Directed graph framework is used and multiple scans with progressively increasing time window width are used for edge construction for maximum likelihood trajectories. The current version of the code was developed for applications in multiphase hydrodynamics, e.g. bubble and particle tracking, and is capable of resolving object motion, merges and splits. Feasible object associations and trajectory graph edge likelihoods are determined using weak mass and momentum conservation laws translated to statistical functions for object properties. The code is compatible with n-dimensional motion with arbitrarily many tracked object properties. This framework is easily extendable beyond the present application by replacing the currently used heuristics with ones more appropriate for the problem at hand. The code is open-source and will be continuously developed further.
	\end{abstract}


	\section{Introduction}
	
		\textit{Multiple Hypotheses Tracking} (MHT) is classically considered to be the most reliable method for finding the optimal solution for data association problems \cite{mht-revisited, mht-deanonimized-main, mht-blackman, mht-cox-efficient-implementation, mht-reid-og}. However, it is often avoided for real-time (online) applications due to its $\mathcal{O}(n)=2^n$ (brute force) computational complexity when searching for optimal and compatible object trajectories, where $n$ is the number of considered associations. There have been attempts to solve the tracking problem by other means, but this always means trading precision/robustness for decreased complexity \cite{mht-revisited, mht-blackman}. While one could reduce the effective $n$ for MHT, this does not affect complexity scaling, which is a problem in cases of large datasets often encountered in scientific applications. In this paper, a new optimization approach is proposed for offline tracking with MHT. The association search problem is formulated as an exact cover problem, which is then solved using \textit{Algorithm-X} \cite{knuthDancingLinks2000b}. This way, the best-case computational complexity becomes $\mathcal{O}(n)=n\log{n}$, allowing to cover the entire search-space of viable solutions, while entirely omitting contradictory solutions.
	
		The tracking algorithm described herein was developed in \textit{Python} for the analysis of multiphase hydrodynamic systems, such as bubble and particle flow in liquids. In our field of research, there are several cases of interest where explicit, precise and robust object tracking is desired: dynamic optical, X-ray and neutron imaging of argon bubble flow in liquid gallium or GaInSn eutectic alloy \cite{birjukovsPhaseBoundaryDynamics2020, birjukovsArgonBubbleFlow2020, megumi-x-rays, megumi-cfd, x-ray-bubble-breakup, x-ray-bubble-coalescence}; neutron imaging of gadolinium oxide particle flow in liquid gallium \cite{neutrons-particles-lappan, neutrons-particles-stirrer-scepanskis, neutrons-particles-stirrer-scepanskis-2, neutrons-simulations-stirrer-valters}; optical imaging of salt crystals and liquid crystal tracers in water \cite{sten-anders-spectral-random-masking, sten-anders-vel-temp-measure}; neutron imaging of gadolinium particles in froth \cite{neutrons-particles-froth-heitcam-tobias}; bubble flow simulations \cite{birjukovsPhaseBoundaryDynamics2020, birjukovsArgonBubbleFlow2020, megumi-cfd}. In all of these cases, the objective is to trace  objects (bubbles/particles segmented beforehand) in time, reconstructing their trajectories, resolving merging and splitting events (i.e. bubble coalescence and particle adhesion) where these occur and performing particle tracking velocimetry (PTV). The output can then be used for an in-depth analysis of bubble/particle collective dynamics, comparison of simulation and experimental data, etc. The code was also designed to be able to account for significant shape oscillations and parameter variations for objects that are typical in the above applications. 
		
	    While this means that the utilized heuristics such as association probability functions and tracing examples showcased herein are mostly applicable to the above cases, the main framework described in this paper is universal and can be readily adapted to any type of offline tracking problem, abstract or physical. Most modern tracking algorithms are designed for use on live data (military, surveillance, transportation and other related applications) \cite{mht-revisited}, whereas here the offline implementation is dictated by scientific use, where real time data interpretation is not required, but rather reliable solutions are expected for potentially very large scale problems within acceptable time intervals (i.e. not necessarily requiring high performance computing).
		
		Directed graph representation for input data and trajectories was chosen because of its intuitive interpretation, which makes the presented tracing framework conceptually easier to implement and customize for the application at hand \cite{mht-deanonimized-main}. This MHT implementation, referred to as MHT-X in this paper, consists of several core components, in this case \textit{Python} objects, where each object represents a stage/method for finding the optimal solution for the tracking problem -- the objects and methods are covered in the following section.

	\section{The algorithm}
	
		\begin{figure}[!h]
			\centering
			\includegraphics[width=1\linewidth]{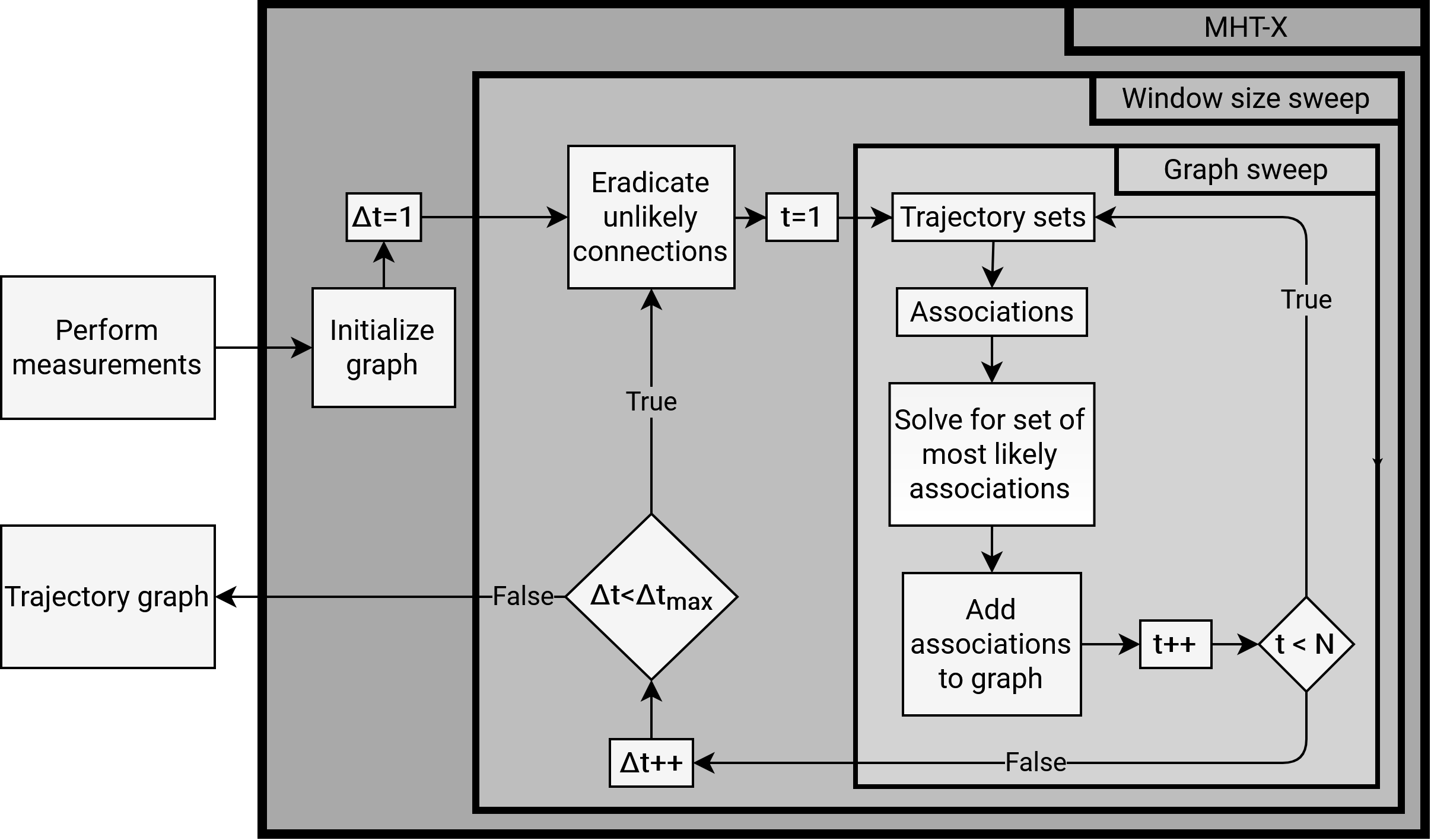}
			\caption{A flowchart of the developed tracing algorithm MHT-X.}
			\label{fig:flowchart}
		\end{figure}
	
		The algorithm starts off with an initialization step, where the data is used to construct an initial graph. The algorithm then proceeds to scan the graph over time with a time window with an iteratively increasing width. This is done to progressively resolve object associations over different time scales.
		
		Each time a new time window width value is set, unlikely edges are eliminated and a graph sweep over time is executed, where at every time step the following operations are performed:
		\begin{enumerate}
			\item Objects (trajectories) with unresolved endpoints are found.
			\item Associations are formed for these objects.
			\item Optimal hypotheses for associations are identified.
			\item The optimal solution is added to the graph.
		\end{enumerate}
		
		The time window width starts with a value of 1 (two neighbouring frames are considered) and is incremented up to a value for which the utilized predictive models of object motion are no longer reliable. A schematic representation of the framework described above can be seen in Figure \ref{fig:flowchart}.

	\subsection{Input and measurements}
		Since the tracking algorithm is offline, a complete set of measurements for object detection events is required as input. For each detection event an object $O_{ti}$, where $t$ is the time index, indicating detection time (frame number) and an $i$ is an auxiliary trajectory index, indicating which trajectory $O_{ti}$ belongs to. It is auxiliary in that it is used herein for mathematical formulation, but is not utilized in the implemented code. $t,i \in \mathbb{N}$ where $t \in [1,N]$ and $N$ is the number of consecutive frames in the input dataset. Each $O_{ti}$ is also prescribed a set of properties, which in this case is $\{ \vec{r}, S \}$ where $\vec{r}$ is the radius vector for the object centroid and $S$ is the projection area due to optical/x-ray/neutron transmission. Generally, however, there are no restrictions on the type and number of properties, but rather this is dictated by the intended application of the present framework.

	\subsection{Directed graph architecture}	
	
		The time-forward directed graph representation of data is the natural framework for MHT and specifically for the problem at hand, and makes the data association formation effortless, since instead of explicitly constructing trajectories, associations are defined between datapoints and then based individual trajectories can be analyzed as necessary. 
		
		A directed graph $G$ is defined by a set of nodes $V$, a set of edges $E$ and a function $P$. The node set consists of the detected objects and two \textit{special nodes} ($V^*$). The special nodes are labeled \textit{Entry} and \textit{Exit}, these are auxiliary (padding) nodes of the graph that represent start and end points of trajectories, respectively. Directed edges of this graph are represented by an unordered set $E$ of ordered pairs of elements of $V$. The function $P$ maps every edge in the graph to its respective likelihood value $p$.

		\begin{equation}
			\begin{cases}
				V = \{O_{ti}\}\cup V^*\\
				E = \{(a, b)\}& a, b \in V, a\not=b, \{a,b\}\not=V^*, a \not=\text{\textit{Exit}}, b\not=\text{\textit{Entry}}\\
				P:E\rightarrow p& p \in(0,1]\\
				G = (V, E, P)
			\end{cases}
		\end{equation}

		Edges represent associations between the nodes and are assigned likelihood values to which $P$ maps every edge. Likelihood represents the probability that the association is true. Node associations represented by directed edges are aligned with the arrow of time. Edges connecting graph nodes to the special nodes are used to store the likelihoods that the connected trajectory nodes are trajectory endpoints. Multiple edges directed away from or towards a node that is not a special node represent split/merge events, respectively, and each of the edges from one such event has an equal likelihood value, which represents the likelihood of that event.

\begin{figure}[!h]
			\centering
			\includegraphics[width=0.75\linewidth]{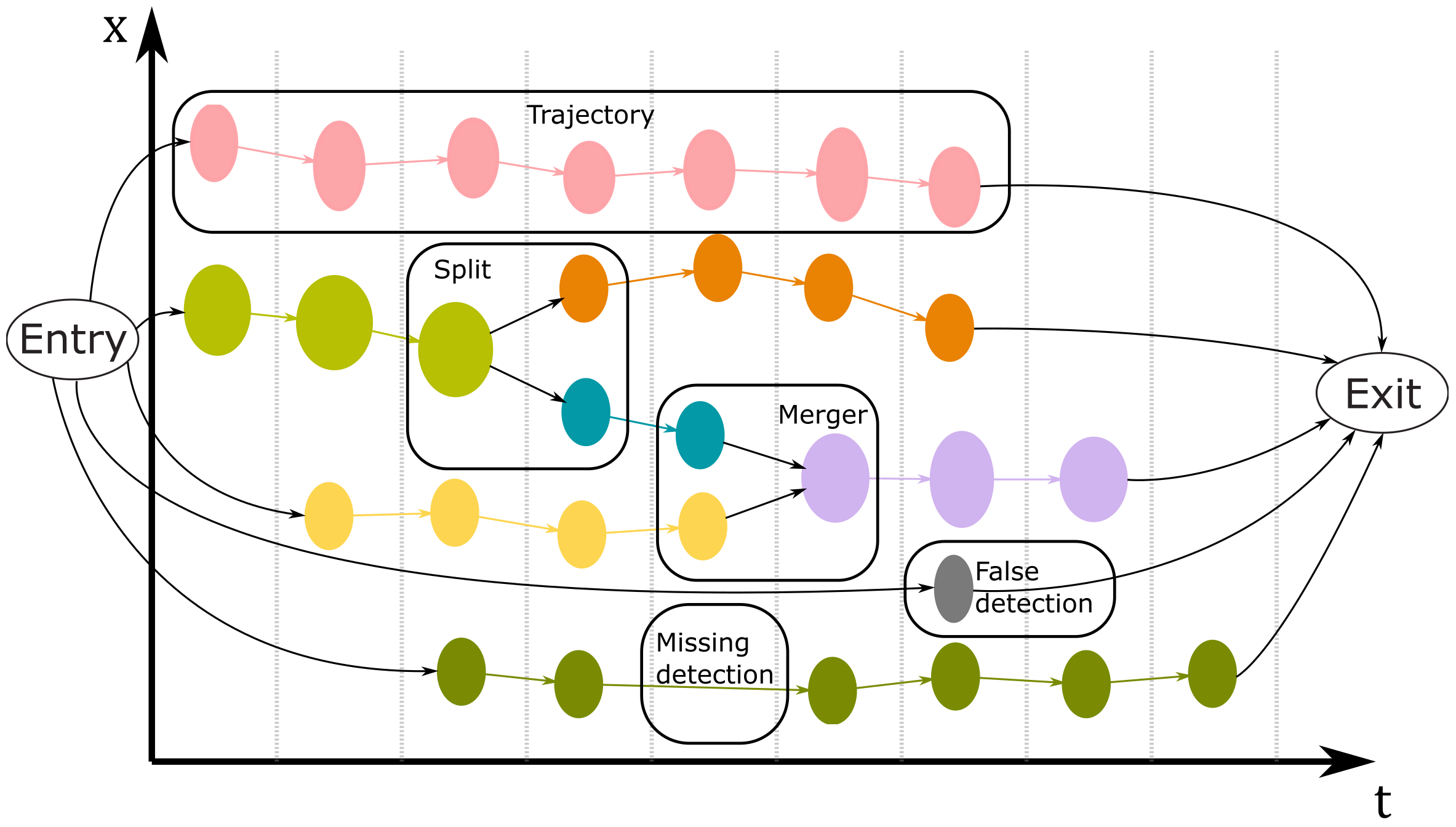}
			\caption{An example directed graph (\textit{trajectory graph}), with different colours signifying separate trajectories.}
			\label{fig:examplegraph}
		\end{figure}

		The directed graph formed by this algorithm represents different trajectories encompassed by it where trajectories are potentially logically connected via split/merge events. A schematic representation of a graph formed after tracing is shown in Figure \ref{fig:examplegraph} and its structure is described in more detail below.

\clearpage

	\subsection{Trajectories}		
		
		A trajectory is a set of sequentially connected objects that were measured over different time frames, which in this framework translates to sets of nodes connected by directed edges, bounded by special nodes and/or split or merge events. In cases where one-to-many split or many-to-one merge events occur, the source trajectories are broken and new ones are formed, generating a set of \textit{logically connected} trajectories -- trajectory \textit{families}. Families are still bounded by special nodes. Nodes which are only connected to the special nodes are considered to be false detections. The latter is not a design feature, but rather is an emerging property of the framework. 
		
		The set of all trajectories is defined as follows:

    	\begin{equation}
    	    T = \left\lbrace \{ T_k \} ~ \bigg\vert ~ T_k = \{ O_{tk} \}, t \in [t_1;t_k], k \in \mathbb{N}  \right\rbrace
    	\end{equation}

        Given a trajectory, one may also define functions that map trajectories to other mathematical objects: this way, one can derive velocity and acceleration over trajectories, as well as use trajectory data to make predictions regarding object motion -- all of this can be used to define heuristics for more efficient object tracking.
        
        To extrapolate a trajectory reliably, a quantitative model is necessary. For MHD multiphase flow, this version of the code utilizes a rather naive general approach for extrapolation: a spline operator $\hat{S}$ is defined, which maps a trajectory to a piece-wise polynomial function $g_k$ of time

        \begin{equation}
        \hat{S}: T_k \rightarrow g_k(t)    
        \end{equation}

        This spline is then used for extrapolation as follows:

        \begin{equation}
            \begin{split}
                T_k(t| t > t_k) = g_k(t_k) + \dv{}{t} ~ g_k(t - a) \cdot (t - t_k) \\
                T_k(t| t < t_1) = g_k(t_1) + \dv{}{t} ~ g_k(t + a) \cdot (t - t_1) \\
            \end{split}
        \end{equation}

    \clearpage

	\subsection{Initialization}
	
		During the initialization step the nodes of $G$ are defined from the set of measurements. A critical step here is to generate boundary conditions (constraints). It is known \textit{a priori} that all nodes with $t = 1$ or $t = N$ must be the endpoints of trajectories. As an initial condition for the graph, all nodes in the graph and the special nodes are interconnected, and each of the edges is prescribed a likelihood value based on the following Boolean statements. The set of edges formed in this step formally is defined as follows:
		
		\begin{equation}
			\begin{split}
			E_1 = \left\lbrace (\text{Entry}, O_{ti}) ~ \bigg\vert ~ \forall O_{ti}, P: (\text{Entry},O_{ti}) \rightarrow
				\begin{cases}
					1, & t = 1\\
					0, & t \not= 1\\
				\end{cases} ~ \right\rbrace\\			
			E_2 = \left\lbrace(O_{ti}, \text{Exit}) ~ \bigg\vert ~ \forall O_{ti}, P: (O_{ti}, \text{Exit}) \rightarrow
				\begin{cases}
					1, & t = N\\
					0, & t \not= N\\
				\end{cases} ~ \right\rbrace\\	
			E = E_1\cup E_2
			\end{split}
		\end{equation}
		
		This step is necessary so as to nodes which lay in either the first or last frame would have unambiguous origins/continuations and ensures that these connections will not be broken at the connection eradication step.

	\subsection{Graph \& time window width sweeps}

		The \textit{graph sweep} is the main part of the algorithm -- it iteratively performs hypotheses definition and evaluation, optimal association search and graph edge insertion using subroutines described later in this section. For this, a time window is defined with time width $\Delta t$, i.e. the number of frames covered by the window is $\Delta t + 1$. This window is then translated forward in time frame-by-frame through the graph inserting new edges within each time window where appropriate. Before this graph sweep is executed, a fully connected graph is expected, i.e. every node (except special nodes) must have incoming and outgoing edges. After this initial graph is formed, a fraction of its edges are removed using the edge eradication method. This yields a sparsely connected graph where some nodes have no incoming and/or outgoing edges. These ambiguities are resolved during the graph sweep yielding a fully connected graph that can be taken as the solution to the tracking problem or can be used as an initial condition for a new graph sweep with a different time window width. 
		
		In the current implementation this is done repeatedly, constituting a \textit{time window width sweep}. After an initial fully connected graph is formed and edge eradication is invoked, $\Delta t = 1$ is set defining a 2-frame window with $t \in [t + \Delta t]$. A graph sweep with this window resolves  the most obvious associations formed between sequential frames. The resulting graph, after unlikely edge eradication, is used as an initial condition for the next graph sweep with a 3-frame interrogation window ($\Delta t = 2$). This process is repeated with increasing $\Delta t$ until it exceeds a user defined threshold. This is done so that consecutive graph sweeps resolve associations over greater distance in time, allowing to overcome detection failures. The following subsections describe the underlying subroutines in detail.

		\subsubsection{Unlikely edge eradication}
		    
		    After every graph sweep and initialization, all measurement nodes will have incoming and outgoing edges. New initial conditions for the next time sweep are generated by removing unlikely edges, which are identified as follows. Using the $p$ distribution for all of the graph edges and a user defined \textit{quantile} parameter $q \in [0,1]$, a likelihood threshold $p_c$ is computed. Edges with $p \leq p_c$ are eradicated.

		\subsubsection{Sets of associable trajectories}
			
			At every time step in the graph sweep for a given $t$ and $\Delta t$ two unordered sets of trajectories are formed: $F_t$ -- a set of trajectories which do not have an endpoint within $[t; t + \Delta t)$; $B_t$ -- a set of trajectories which do not have an endpoint within $(t; t + \Delta t]$. Formally $F_t$ and $B_t$ are defined as
			
			\begin{equation}
				F_t = \left\lbrace \{ T_k \} \bigg\vert t_k \in [t; t+\Delta t) \land \nexists (O_{t_kk}, \forall O_{\hat{t} i}) \right\rbrace
			\end{equation}
			\begin{equation}
				B_t = \left\lbrace \{ T_k \} \bigg\vert t_1 \in (t; t+\Delta t] \land \nexists (\forall O_{\hat{t} i}, O_{t_1k}) \right\rbrace
			\end{equation}
			where $\hat{t} \in [1;N]$.

		\subsubsection{Associations}
			
			The types of associations $A$ supported by the current version of the tracer are formally defined as follows:
			
			\begin{equation}
				\begin{cases}
					\text{Entry} 		& (\varnothing, T_k), T_k \in B_t\\
					\text{Exit}			& (T_k, \varnothing), T_k \in F_t\\
					\text{Translation}	& (T_k, T_m), T_k \in F_t \land T_m \in B_t\\
					\text{Split}		& (T_k, \{T_m\}), T_k \in F_t \land \forall T_m \in B_t\\	
					\text{Merge}		& (\{T_k\}, T_m), \forall T_k \in F_t \land T_m \in B_t
				\end{cases}
			\end{equation}

			The \textit{associator} routine in the code takes an element of $F_t$ and using the pairwise association condition determines which of the elements of $B_t$ can be associated with it. This results in a set of elements associable to $F_t$. Because each $F_t$ element can be associated to many elements from $B_t$, an adjustable constraint $\gamma \in \mathbb{N}_0$ is introduced to limit the number of possible split/merge components associated with a given element. In the limit case of $\gamma = 0$ none of the associated $F_t$ and $B_t$ elements are taken and an empty set is associated. The same procedure is then repeated for the elements in $B_t$ except 1-to-1 associations are not considered, since these are already covered in the first association scan.
				
			In the current implementation, before the trajectory association problem is formed, it is checked if associations satisfy pairwise association conditions and association constraints that are outlined in the following sections.	If not, the violating associations are removed. This is important to reduce the large number of pairwise associations to a set of a manageable size.
			
            While here we consider associations defined between trajectories, in the trajectory graph they are represented by edges between the endpoints of trajectories. Associations are mapped to edges via the following map $K$:
            
            \begin{equation}
            \label{eq:association-map}
            K:A\rightarrow E =
                \begin{cases}
                (\varnothing, T_k)&\rightarrow (\text{Entry}, O_{t_1k})\\
                (T_k, \varnothing)&\rightarrow (O_{t_kk}, \text{Exit})\\
                (T_k, T_m)&\rightarrow (O_{t_kk}, O_{t_1m}), ~ k \neq m, ~ m \in \mathbb{N}\\
                (T_k, \{T_m\})&\rightarrow \{(O_{t_kk}, O_{t_1m})\}, ~ k \neq m, ~ m \in \mathbb{N}\\
                (\{T_k\}, T_m)&\rightarrow \{(O_{t_kk}, O_{t_1m})\}, ~ k \neq m, ~ m \in \mathbb{N} 
                \end{cases}
            \end{equation}

			\subsection{Association conditions}

        		\begin{itemize}		
        		
                    \item Self-associations are forbidden.
                    \item Time-forward associations only.
                    \item Limited maximum object displacement per frame.
                    \item Limited association range:
        
                    \begin{itemize}

                        \item A primary sphere of influence (SOI) based on the node object's effective radius\\
                        
                        \begin{equation}
                            |\vec{r_0} - \vec{r_k}| < C \cdot r_{SOI}(S)
                        \end{equation}
                        where $C$ is a control parameter.
                        
                        \item A secondary smaller SOI of a fixed size -- objects within are always associated.
                    
                    \end{itemize}		
                		
                \end{itemize}

			\subsection{Association constraints}
			
                Association constraints determine whether or not an association is plausible. Entry and exit associations are always considered plausible. Translation associations are expected to comply with weak mass and momentum conservation laws, which are used to determine if two trajectories are consistent in terms of object motion.

                Denote the two trajectory segments within the time window with subscripts 1 and 2 and the connecting edge with subscript $k$. Translation associations are constrained by the maximum linear acceleration $a_c$:

                \begin{equation}
                \begin{split}
                    2 \cdot \frac{\|\vec{v}_k - \vec{v}_1\|}{\Delta t_k + \Delta t_1} < a_c \\
                    2 \cdot \frac{\|\vec{v}_2-\vec{v}_1\|}{\Delta t_2 + \Delta t_k} < a_c
                \end{split}
                \end{equation}
    			where $\vec{v}$ is velocity at the respective trajectory edges and $\Delta t$ is the time difference between the nodes of considered edges.
    			
    			We also limit the change in direction of movement based on the velocity:
    			\begin{equation}
    			\begin{split}
    			    \arccos\left(\frac{\vec{v}_k\cdot\vec{v}_1}{\|\vec{v}_k\|\cdot\|\vec{v}_1\|}\right) < (\pi + \epsilon) \cdot \exp \left( -\frac{\|\vec{v}_1\|}{\lambda} \right) \\
    			    \arccos\left(\frac{\vec{v}_2\cdot\vec{v}_k}{\|\vec{v}_2\|\cdot\|\vec{v}_k\|}\right) < (\pi + \epsilon) \cdot  \exp \left( -\frac{\|\vec{v}_k\|}{\lambda} \right)
    			\end{split}
    			\end{equation}
    			where $\epsilon$ is an arbitrary small constant, and $\lambda$ is a control parameter. The direction deviation constraints mimic momentum conservation in that it is expected that objects with greater velocity are less susceptible to deflection.

    			Weak mass conservation limits projection area differences between trajectories connected via translation:

    			\begin{equation}
    			    T_1, T_2: \frac{| \left< S_1 \right> - \left< S_2 \right> |}{\max \left( \left< S_1 \right>, \left< S_2 \right> \right) } < \varepsilon_t \cdot \left< \frac{\sigma_k}{ \left< S_k \right>} \right>
    			    \label{eq:constraint-translation}
    			\end{equation}
    			where $S_{1,2}$ are the sets of area measurements, $\sigma_{1,2}$ are the standard deviations for $S_{1,2}$, $k=\{1,2\}$ and $\varepsilon_t$ is the association threshold.

    			In the case of split/merge events, weak momentum conservation is unreliable due to surface tension effects, therefore only weak mass conservation is used. This essentially checks if the projection areas of objects before and after splits/merges are consistent.

    			\begin{equation}
    			    \begin{split}
    			    \frac{ | S_0 - \sum_k \left< S_k \right> | }{ \max \left( S_0, \sum_k \left< S_k \right> \right) } < 
    			\varepsilon_s \cdot \left< \frac{\sigma_k}{ \left< S_k \right>} \right> \\
    			    \end{split}
    			    \label{eq:constraint-interaction}
    			\end{equation}
    			where $S_k$ and $\sigma_k$, $k \in \mathbb{N}$ correspond to the merge components/split products and $\varepsilon_s$ is the association threshold.

		\subsection{MHT}

			The Bayesian formulation of MHT is used \cite{mht-deanonimized-main}. Due to the offline nature of the algorithm, the framework is greatly simplified and the problem of finding feasible (non-contradictory) sets of trajectories is as follows. Let $A$ be a list of feasible associations and $X$ be a binary vector representing hypothesized trajectory configurations (i.e. sets of association states). The goal is to find the most likely state from $X$ given $A$, that is, to find the maximum \textit{a posteriori} estimate $X^*$:
		
			\begin{equation}
				X^* = \arg\max\limits_Xp(X|A)
			\end{equation}
where due to the Bayes' theorem one has
	
			\begin{equation}
				p(X|A) \propto p(A|X)p(X)
			\end{equation}
where $p(X) = 1$ if the hypothesized associations are not contradictory, meaning that a trajectory is in only one of the hypothesized associations. Given a trajectory configuration, one calculates the likelihood of associations $A_i$ as follows:

			\begin{equation}
				p(A|X)=\prod_{i=0}^{n}p(A_i|X_i)
			\end{equation}
			\begin{equation}
				p(A_i|X_i) = 
				\begin{cases}
					f(A_i), 		&  ~ X_i = 1\\
					1 - f(A_i), 	&  ~ X_i = 0
				\end{cases}
			\end{equation}
            where $f(A_i)$ is a function that evaluates the likelihood of a single association assumed to be \textit{True}. Herein $f(A_i)$ are referred to as \textit{statistical functions}. A unique statistical function is defined for every type of association. 

			The trivial way to solve for $X^*$ is a brute force search verifying every possible $X$, which is extremely inefficient due to the $\mathcal{O}(n)=2^n$ complexity of the search, and is therefore only feasible for a very low number of associations. Reducing the effective $n$ is an option, but that does not solve the scaling problem which becomes critical for very large measurement sets, i.e. measurements with high number density per frame or very long measurement processes, which is the case in many scientific applications.
			
			Reduction of the effective $n$ was attempted by representing the set of associations as binary matrix ($A_{ij}$) where each row stands for an element of $F_t$ or $B_t$ and columns represent the individual associations. Then an undetermined system of linear equations was formed: $A_{ij} X_j=b_i$, where $b_i$ is a vector containing only unity elements. Solving this for the independent variables, they could be then used to solve a problem with a reduced $n$. While this method yielded true solutions in many cases and reduced complexity, it sometimes failed to yield any solutions and was therefore unreliable.
			
			The proposed approach is to recognize this as an exact cover problem \cite{knuthDancingLinks2000b}, since the solutions of the exact cover problem are by definition sets of associations that yield $p(X)=1$. The problem is reformulated as an exact cover problem as follows:
			
			\begin{enumerate}
				\item Define the universe as $U = F_n \cup B_m$
				\item $A$ is a collection of subsets of $U$
				\item Solve for collections of elements of $A$ in compliance with the exact cover problem.
			\end{enumerate}
			
			The best known way of solving an exact cover problem is using the \textit{Algorithm X} (Knuth's algorithm), which reduces the computational complexity significantly down to $\mathcal{O}(n)=n\log{n}$ (best case scenario). Complexity is reduced further by clustering associations into disjoint sets before formulating and solving the exact cover problem \cite{disjoint-graphs}.
			
			\subsubsection{Disjoint sets of associations}
			
    			Before formulating the exact cover problem, it is possible to further reduce its complexity by separating the associations into disjoint sets. 
			
			    This is achieved by defining an undirected graph ($G^*$) as follows:
			    \begin{equation}
    			    \begin{cases}
    			        V^* = \{O_{t_kk}| \forall T_k \in F_t\} \cup \{O_{t_1k}| \forall T_k \in B_t\} \\
    			        E^* = \{K(A_i)|\forall A_i\in A\}\\
    			        G^*=(V^*,E^*)
    			    \end{cases}
			    \end{equation}
			
			    A set of disjoint subgraphs is formed from $G^*$ and then by checking which of the subgraphs each association belongs to, one can form the disjoint sets of associations. This very effectively reduces the size of the universe for the exact cover problem.

			\subsubsection{Statistical functions}
			
			Let $\mathcal{N} (x, \mu, \sigma)$ be a Gaussian distribution with its mean $\mu$ and standard deviation $\sigma$. The propagation probability is measured by:
			
			\begin{equation}
			    f_1 = \alpha \cdot 
			    \frac{ \mathcal{N} \left( \delta r, 0, \sigma_{\delta r} \right) }{ \mathcal{N} \left( 0, 0, \sigma_{\delta r} \right) } + (1-\alpha) \cdot 
			    \frac{ \mathcal{N} \left( \delta S, 0, \sigma_{\delta S} \right) }{ \mathcal{N} \left( 0, 0, \sigma_{\delta S} \right) }
			    \label{eq:likelihood-translation}
			\end{equation}
			where $\delta r$ is the displacement magnitude, $\delta S$ is the area difference between node objects and $\alpha$ is the weight adjustment parameter. $\sigma_{\delta r}$ and $\sigma_{\delta S}$ are computed for both $T_k$ considered for connection via a translation edge.
			
			Entry/exit edge probability is:
			
			\begin{equation}
			    f_2 = \frac{1}{ 1 + \exp \left( a (y-b) \right) }
			\end{equation}
			where $y$ is the vertical coordinate in a 2- or 3-dimensional image and $a$, $b$ are control parameters.
			
			The merge/split event probability is computed as follows:
			
	     	\begin{equation}
		    f_3 = \beta \cdot \mathcal{N} \left( \delta (S_0, S_k), 0, \left< \sigma_{S_k} \right> \right) +
		    (1-\beta) \cdot \frac{M_s}{M} \cdot \mathcal{N} \left( \left< \vec{r}_k \right> - \vec{r}_0, 0, \sigma_{\delta r} \right)
		    \label{eq:likelihood-interaction}
		    \end{equation}
			where $\beta$ is the weight adjustment parameter, $S_0$ is the area of the split source/merge product, $S_k$ are the areas of the merge/split components, $\vec{r}_0$ is the position of the split source/merge product, $\vec{r}_k$ are the positions of the merge/split components, $M$ and $M_s$ are the number of involved trajectories and the number of trajectories with available motion prediction (i.e. there are enough points in a trajectory), respectively.

		\subsubsection{Filtering associations}

			It is not always desirable to immediately add the feasible associations (edges) to the graph since new trajectories might be generated or enter the time window in the next iteration of the time sweep that are better solutions to the problem. Therefore any association considered likely (and the resulting trajectories) must also satisfy tree conditions:
			
			\begin{itemize}
                \item $\forall T_k \in F_t : t_k = t$
                \item $\forall T_k \in B_t : t_1 = t + 1$
                \item $p_k > p_c$
            \end{itemize}
            
            This is the last step of the time sweep. After this, the set of accepted edges is added to the graph $G$ via (\ref{eq:association-map}).

		\section{Benchmarks}
		
		To assess the performance of MHT-X for scientific applications, we applied it to three cases of bubble flow in liquid metal where object tracking is necessary and offline tracking is appropriate:
		
		\begin{itemize}
		    \item 2D simulations of bubble flow in a rectangular vessel
		    \item Dynamic X-ray radiography of bubble flow in a rectangular vessel filled with GaInSn wherein bubbles are injected via a top submerged lance (TSL)
		    \item Dynamic neutron radiography of argon bubble chain flow in a rectangular liquid gallium vessel -- bubbles are injected at the vessel bottom via a horizontal/vertical tube
		\end{itemize}
		
		In all three cases segmentation is performed prior to tracing and bubble flow regime is such that bubbles are expected to deform considerably while ascending to the free surface of liquid metal at the top of the vessels.

		\subsection{2D bubble flow simulation}
		   
\begin{figure}[!h]
		        \centering			         \includegraphics[width=1\linewidth]{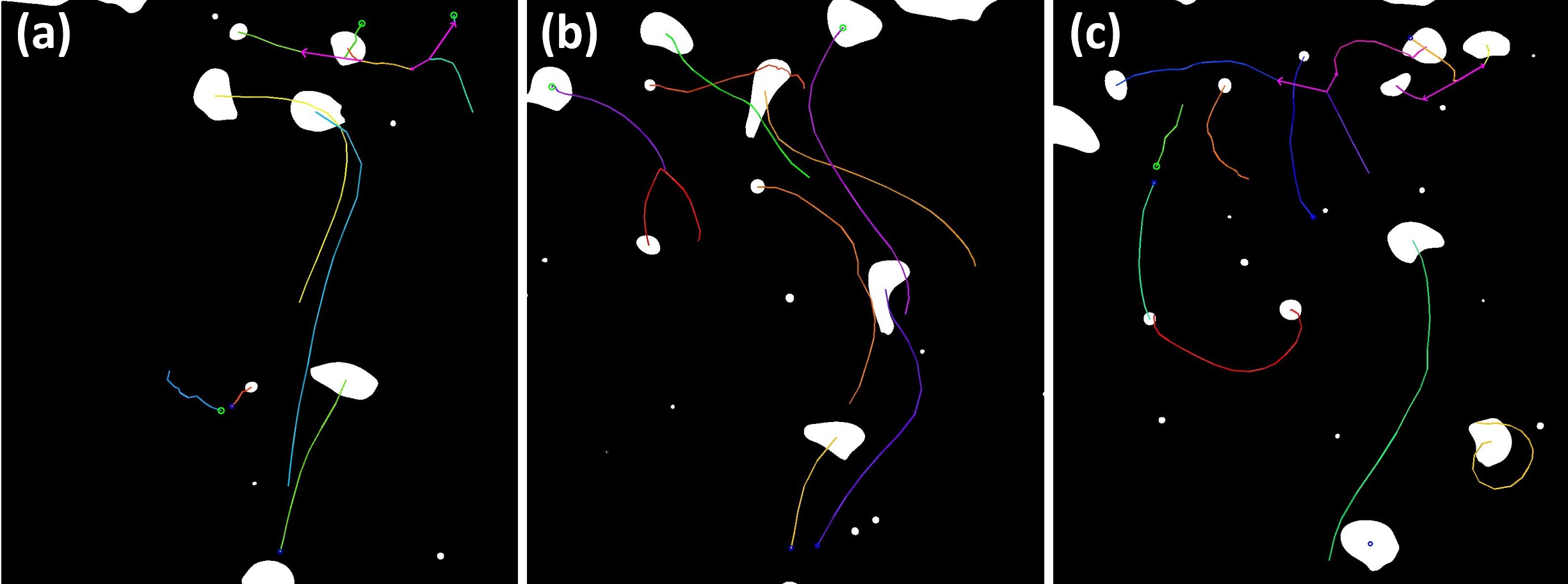}
    			\caption{An example of characteristic trajectory patterns and (c) two splits (purple arrows) in rapid succession resolved in presence of strong deformations and other potentially interfering bubbles. Trajectories are color coded by their IDs.}
    			\label{fig:benchmark-frames}
			\end{figure}

\begin{figure}[!h]
		        \centering			         \includegraphics[width=1\linewidth]{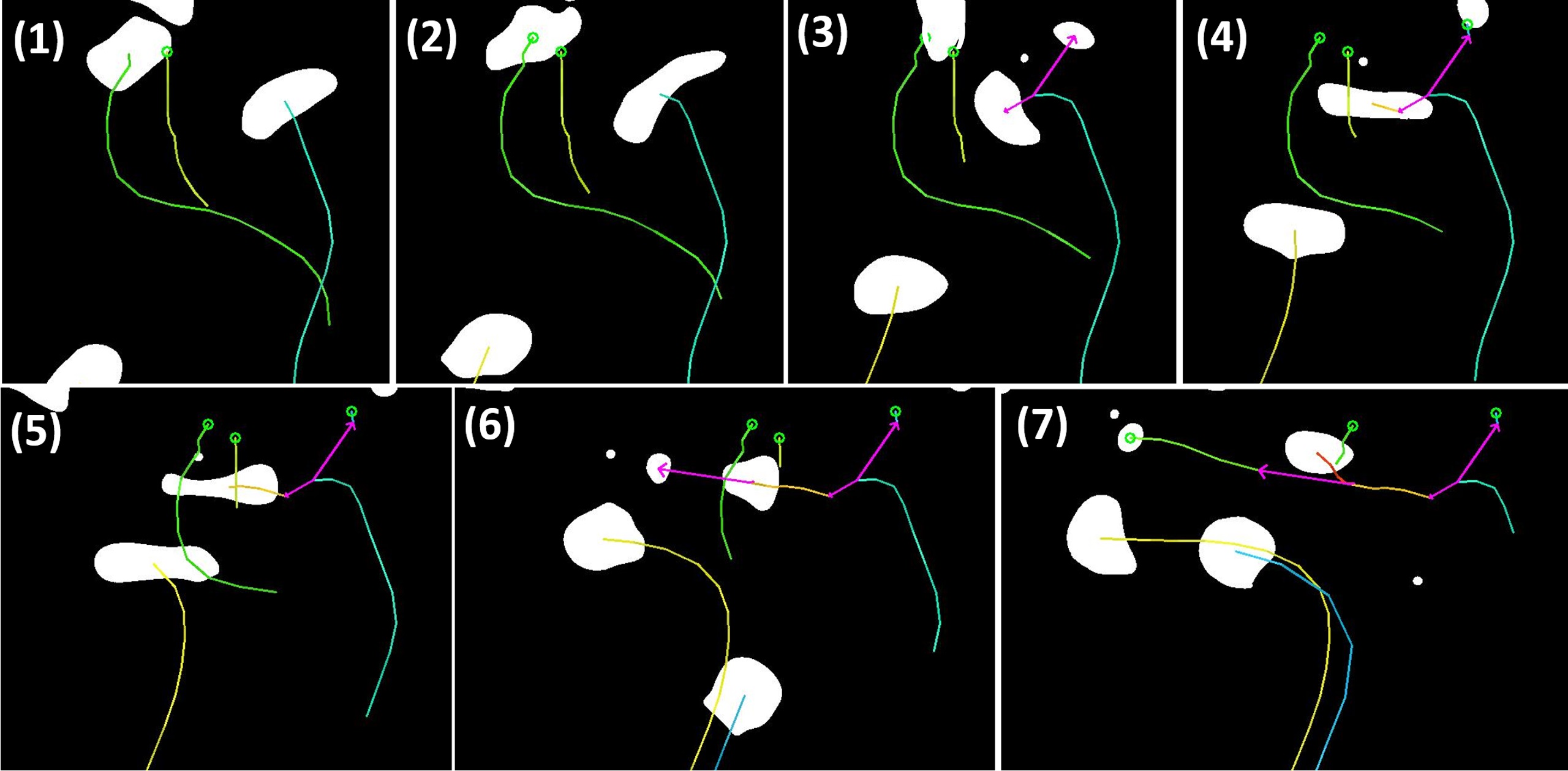}
    			\caption{(1-7) Several nearby bubble trajectories and splitting events resolved over sequential frames from simulation data, and (a,b) examples of characteristic trajectory families.}
    			\label{fig:benchmark-objects}
			\end{figure}

\begin{figure}[!h]
		        \centering			         \includegraphics[width=0.95\linewidth]{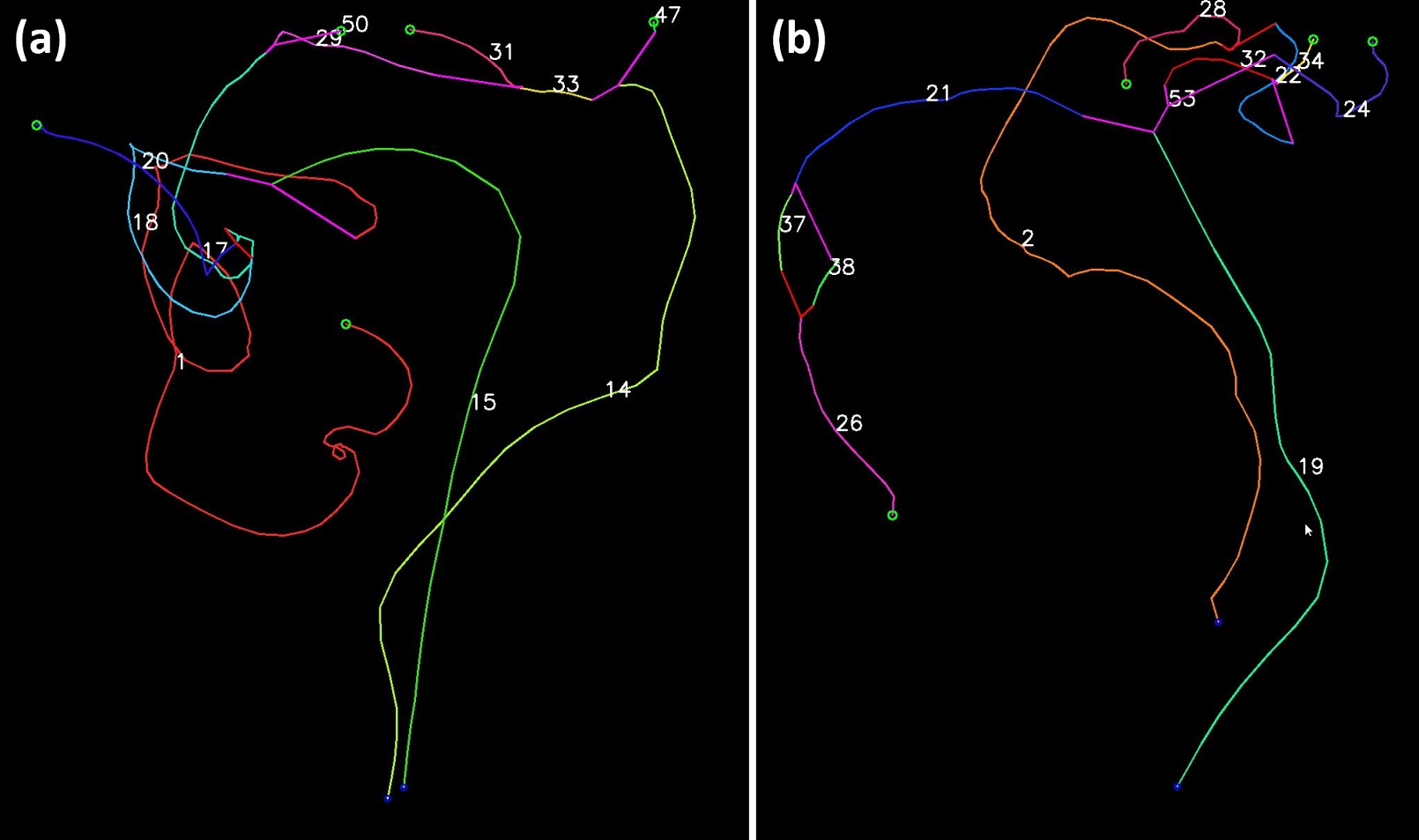}
    			\caption{Two examples of trajectory families recovered from the trajectory graph produced by MHT-X with trajectories color coded by IDs (highlighted).}
    			\label{fig:benchmark-families}
			\end{figure}

\begin{figure}[!h]
		        \centering			         \includegraphics[width=0.75\linewidth]{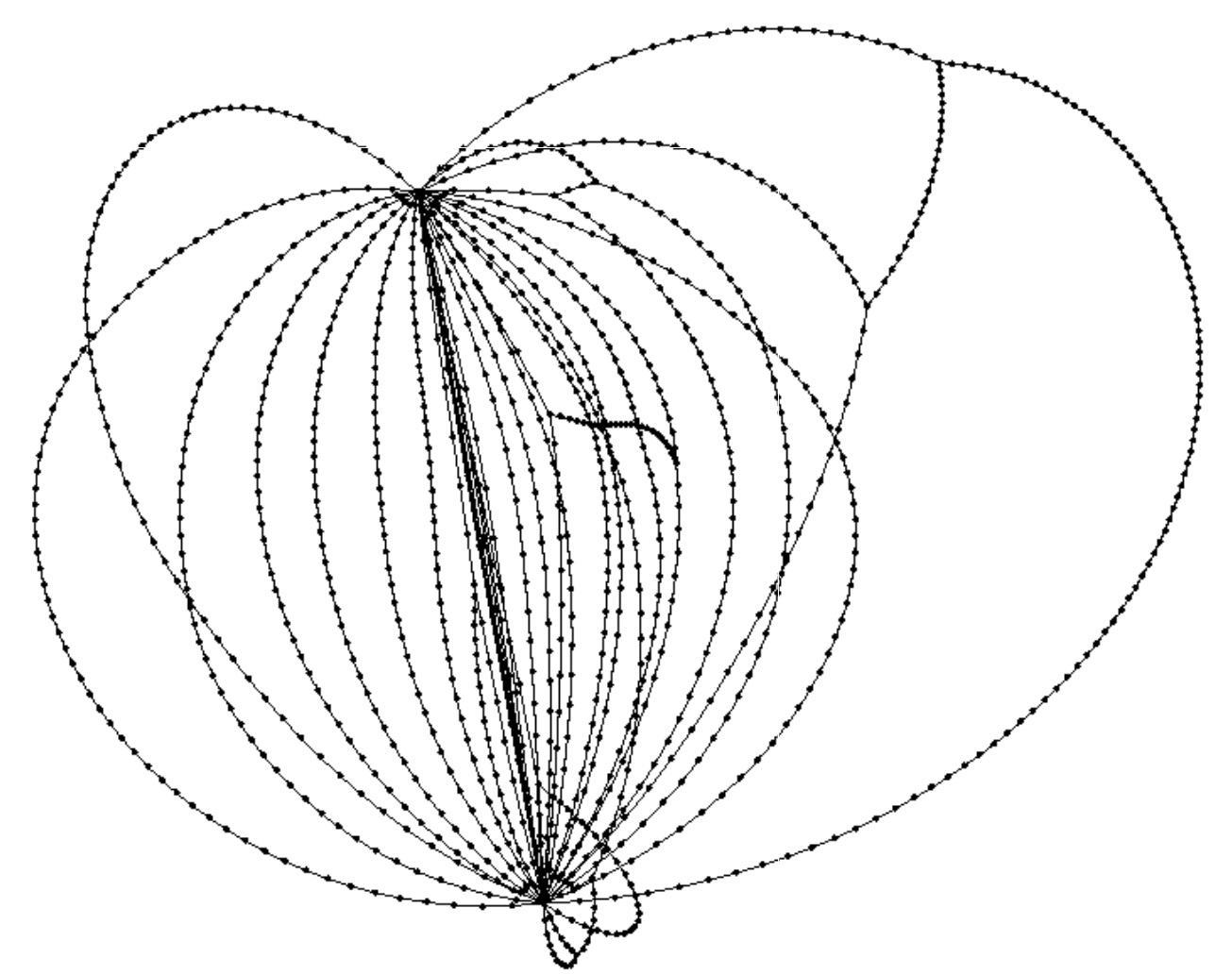}
    			\caption{A solution graph computed by MHT-X for the bubble flow simulation benchmark. This spatially sparse representation of the trajectory graph was rendered in \textit{GePhi}. All trajectories here stem from the entry node (bottom) and converge at the exit node (top). Note that while the coordinates are distorted, the relative vertical (entry to exit direction) positioning is still representative of the actual measurement points (i.e. the relative positions of split/merge events).}
    			\label{fig:benchmark-graph}
			\end{figure}

		   The first benchmark is the output of a 2-dimensional simulation of bubbles rising through liquid gallium in a rectangular vessel. The flow regime has been adjusted such that bubble trajectories are highly irregular and collisions/splits/merges are frequent. The bubbles have been perfectly segmented in that there are no false positives or detection failures in this case. Bubbles with projection areas below a predefined threshold were not measured. Therefore this is an idealized test of the tracking capabilities of MHT-X in case of a rather high number density of objects of various sizes and variable shapes with frequent interactions. Here, object coordinates and projection area are tracked. Examples of the tracing output are shown in Figures \ref{fig:benchmark-frames}-\ref{fig:benchmark-graph}.

		   \clearpage

		   Several frames with overlaid reconstructed bubble trajectories are shown in Figure \ref{fig:benchmark-frames}, where trajectories are color coded by bubble ID. The code successfully tracks both large, significantly deforming bubbles, and smaller ones, even in cases of close proximity with co-linear motion. Note the two detected bubble splits indicated in Figure \ref{fig:benchmark-frames}c with purple arrows. Another important feature of this case is that bubbles move in a variety of patterns: ascension due to buoyancy, slow oscillatory motion due to entrapment in low velocity zones, downward motion due to a large vortex with counter-clockwise mass flow -- all of this is representative of realistic flow conditions in two-phase systems.

			In Figure \ref{fig:benchmark-objects} one can see several bubbles tracked across consecutive frames. Note that in Figures \ref{fig:benchmark-objects}(3)-\ref{fig:benchmark-objects}(6) two split events is rapid succession are resolved where bubble shapes before and after breakup are radically different. It is also important to note that MHT-X does not lose track of bubbles despite significant elongation (especially in frames 3-5) and proximity of two more bubbles that, while initially ascending, divert to the left and begin enter almost co-linear motion (frames 5-7).

		    	Figure \ref{fig:benchmark-families} shows two examples of logically connected trajectory sets (families) derived from the established trajectory graph. Note especially Figure \ref{fig:benchmark-families}a where family members exhibit rather complex trajectories, proximate and even overlapping trajectories. The entire family in Figure \ref{fig:benchmark-families}a originates from a common entry point at the bottom of the field of view (FOV). The graph allows to directly examine the entire trajectory network and qualitatively assess the intensity of collective dynamics and where the interaction events are localized within the FOV.

			In addition to visual information regarding bubble motion and interactions, families and the corresponding exported timestamped datasets for further processing (velocimetry, trajectory curvature measurements, shape parameter evolution tracking, etc.), it may also be helpful to visualize the constructed trajectory graph itself -- the solution graph for this benchmark is shown in Figure \ref{fig:benchmark-graph}.

			However, even though this benchmark demonstrates successful tracing for rather complicated flow patterns and bubble interactions, it is somewhat idealized. Bubble projection area conservation is not violated too strongly, i.e. bubbles do not physically vary in volume (the FOV is above the growth region at the inlet), and the segmented dataset is virtually without error or noise. The following two benchmarks address these conditions.

			\clearpage

		\subsection{Dynamic X-ray radiography of bubble flow}

		The second benchmark stems from a dynamic X-ray imaging of bubble flow in a rectangular vessel with a TSL setup \cite{megumi-x-rays} where the bubbles are injected within the FOV. This means that bubble volume, and therefore also the projection are due to X-ray transmission, are generally very different between a bubble that is being ejected from the TSL tip and already detached bubbles. In addition, the bubbles also exhibit substantial deformations including out-of-plane (with respect to the FOV) motion of the argon/GaInSn interface. Moreover, the bubbles are segmented from X-ray images where noise and potentially artefacts are present, therefore bubble shapes are generally not recovered perfectly. The flow regime here is such that bubbles are expected to interact frequently, even though their number density is less than in the previous benchmark. 
		
\begin{figure}[!h]
		        \centering			         
		        \includegraphics[width=1\linewidth]{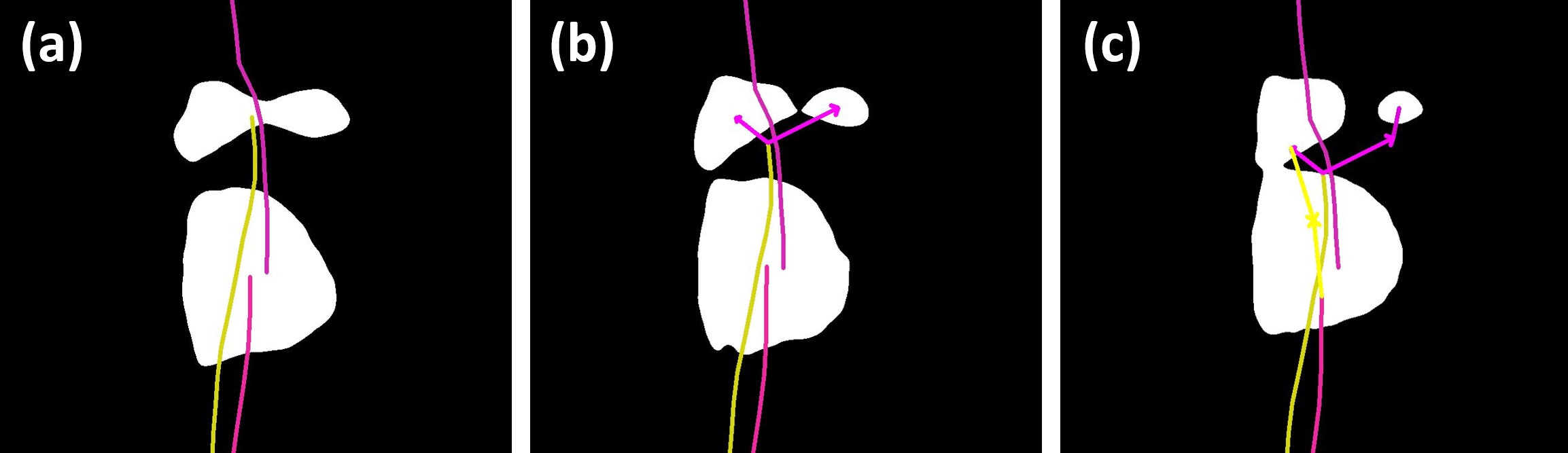}
    			\caption{An instance of a correctly resolved temporally proximate split/merge sequence.}
    			\label{fig:tsl-split-merge}
			\end{figure}

\begin{figure}[!h]
		        \centering			         
		        \includegraphics[width=0.80\linewidth]{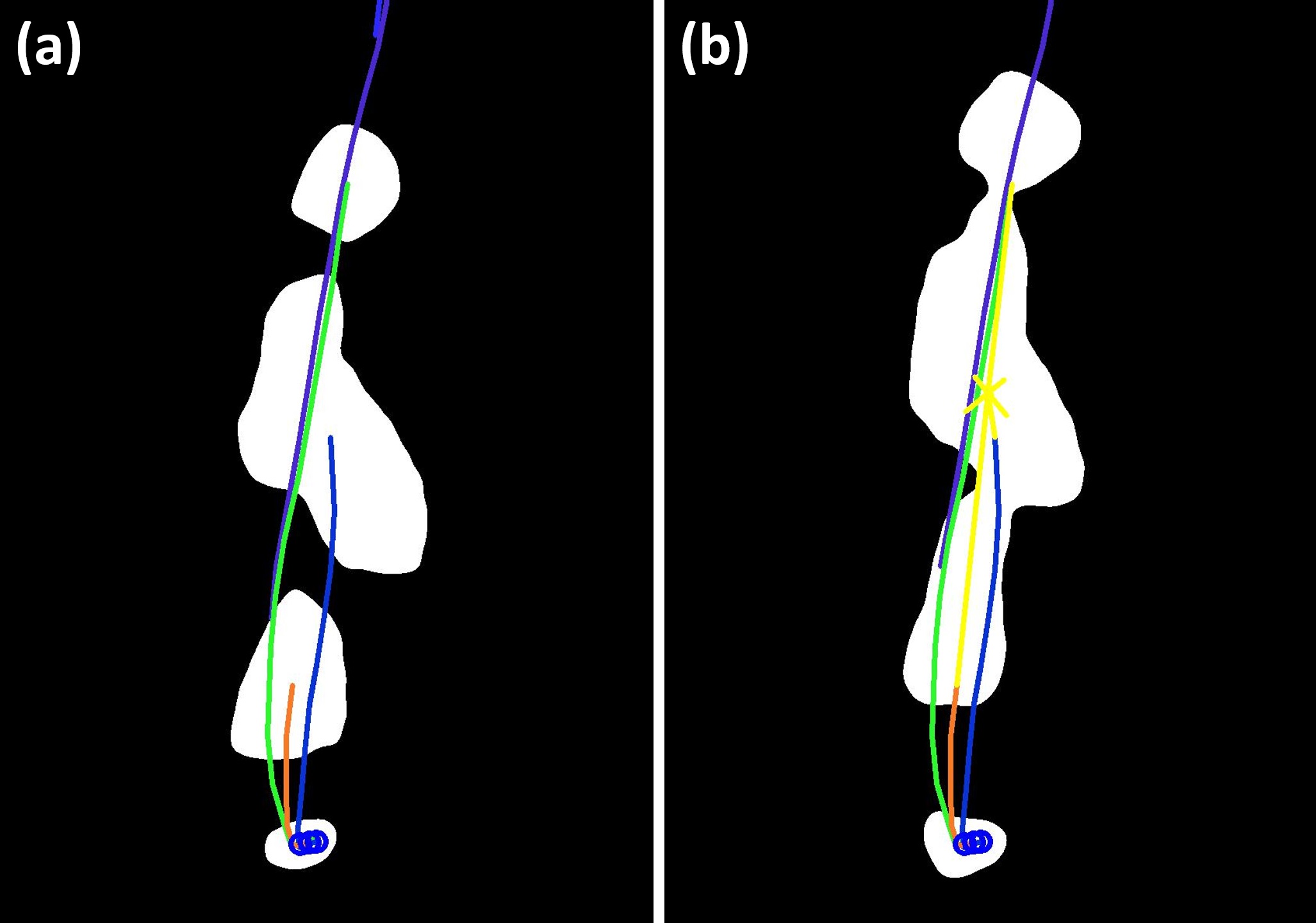}
    			\caption{A three-to-one merge event (yellow arrows) resolved by MHT-X.}
    			\label{fig:tsl-tri-merge}
			\end{figure}

\begin{figure}[!h]
		        \centering			         
		        \includegraphics[width=0.70\linewidth]{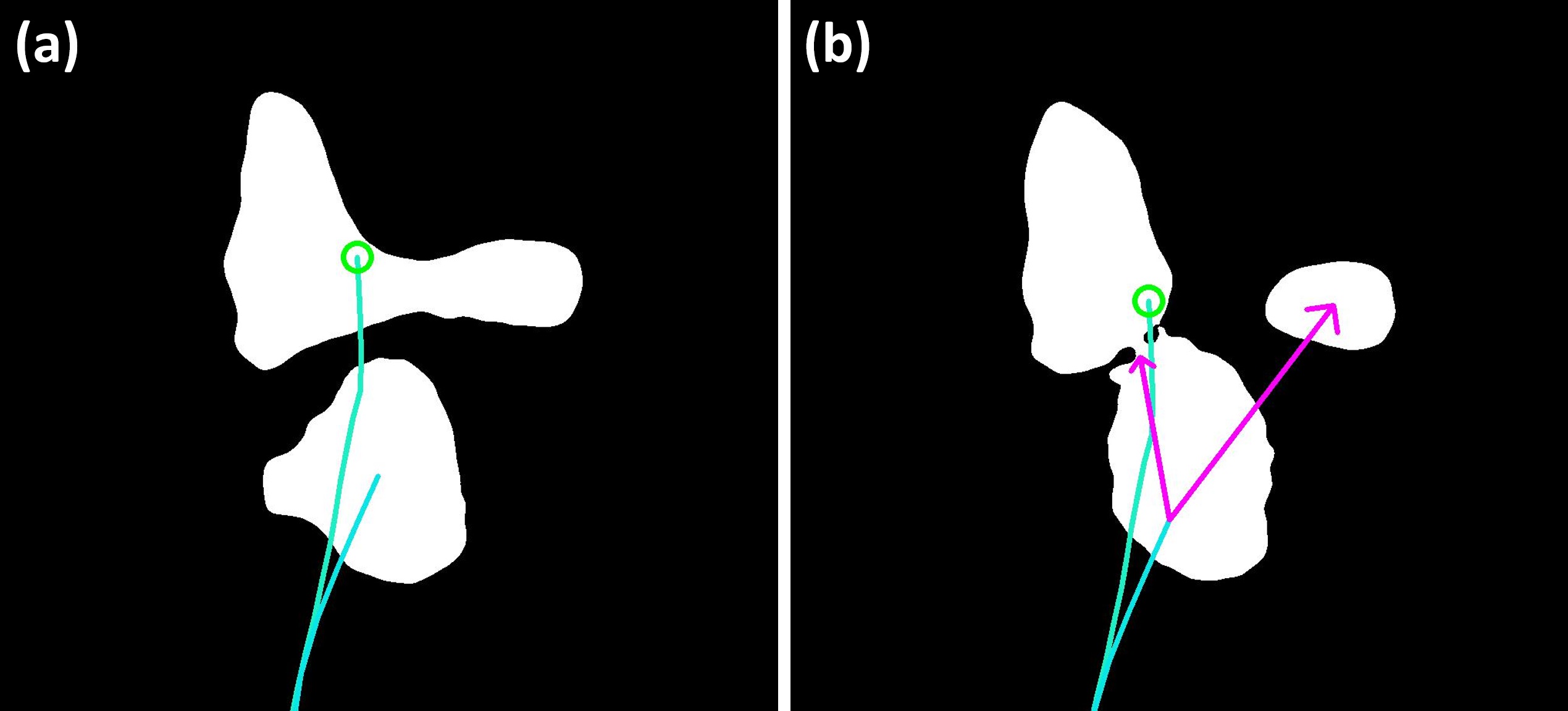}
    			\caption{An instance of a sub-resolution split-merge sequence, a many-to-many event in the context of the input data and the current version of MHT-X.}
    			\label{fig:tsl-splerge}
			\end{figure}

\begin{figure}[!h]
		        \centering			         
		        \includegraphics[width=0.80\linewidth]{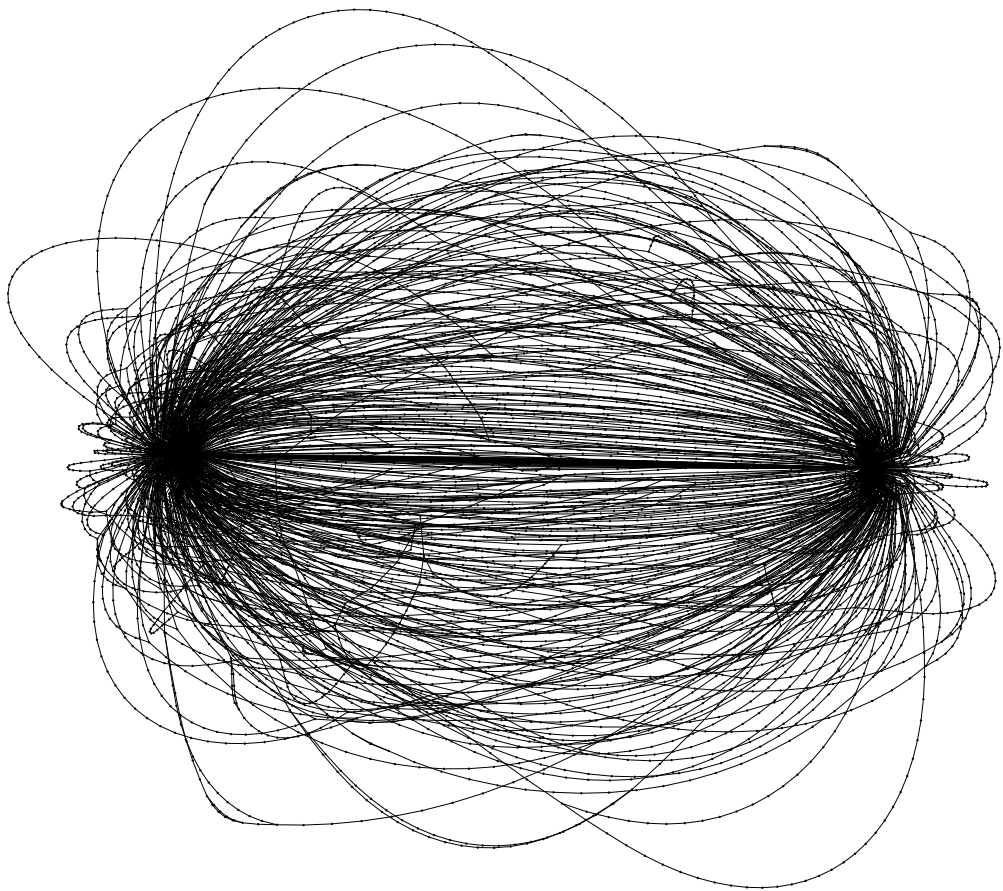}
    			\caption{A solution graph computed for the second benchmark -- the entry node is on the left and the exit node is on the right.}
    			\label{fig:tsl-grapth-all}
			\end{figure}

\begin{figure}[!h]
		        \centering			         
		        \includegraphics[width=1.0\linewidth]{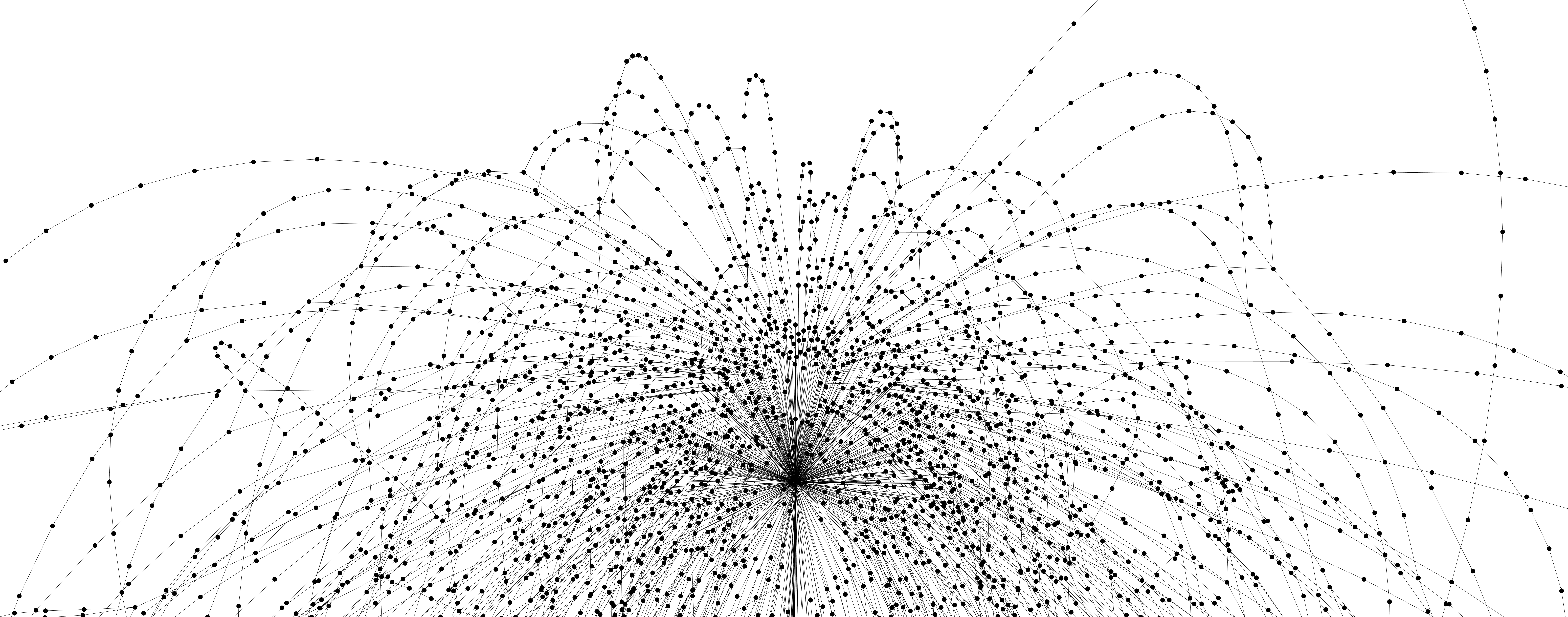}
    			\caption{A close-up view of the entry note region from Figure \ref{fig:tsl-grapth-all}.}
    			\label{fig:tsl-graph-entry}
			\end{figure}

\begin{figure}[!h]
		        \centering			         
		        \includegraphics[width=0.9\linewidth]{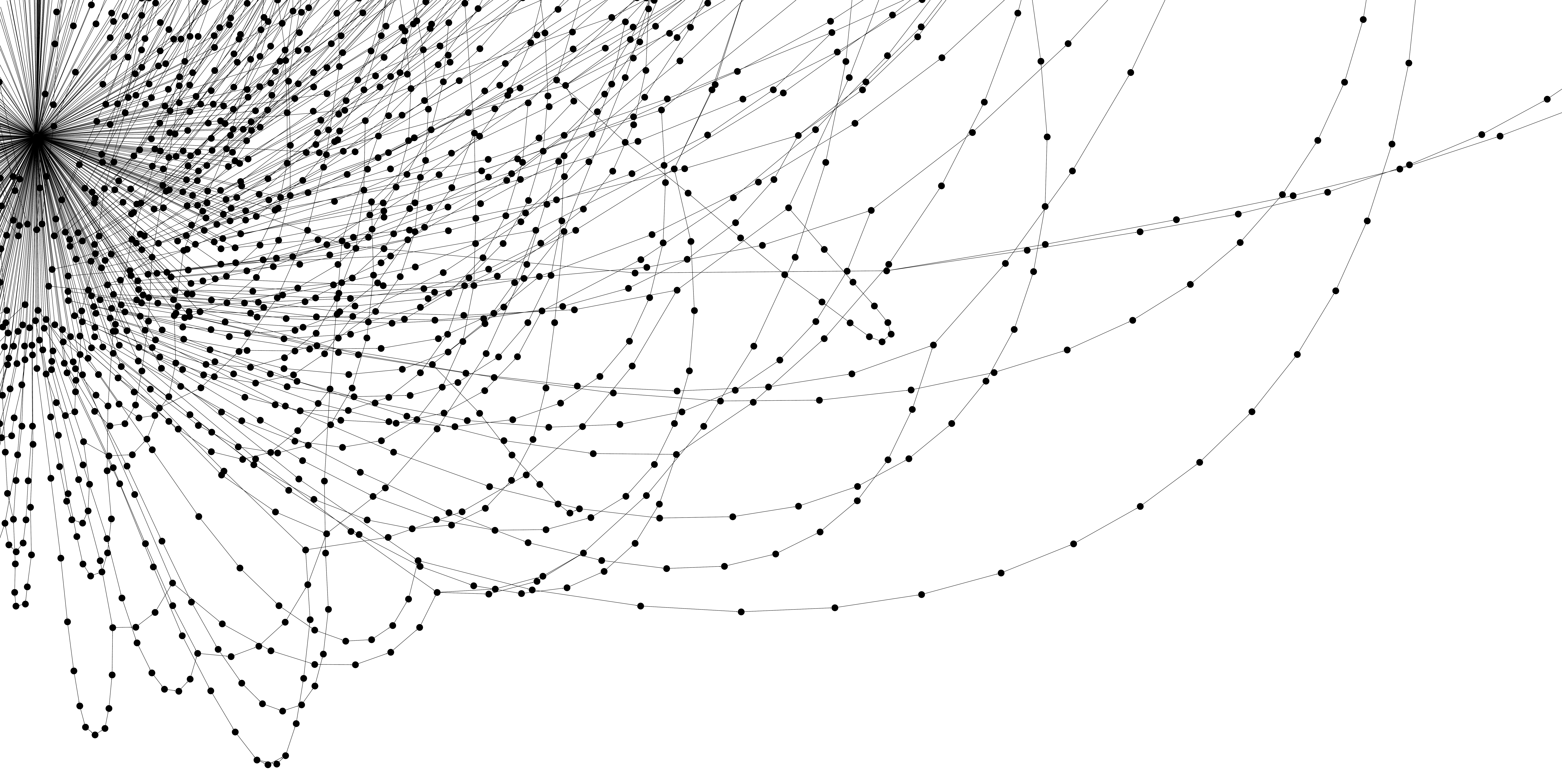}
    			\caption{Another in-detail view of the coalescence events near the entry note region from Figure \ref{fig:tsl-grapth-all}.}
    			\label{fig:tsl-graph-entry-side}
			\end{figure}

		However, in this case the data regarding local volume fraction over the FOV are available from measurements \cite{megumi-x-rays} and bubble volume has been derived and supplied to the tracer. A conservation law of the same form as one for the projection area $S$ in (\ref{eq:constraint-translation}), (\ref{eq:constraint-interaction}),  (\ref{eq:likelihood-translation}) and (\ref{eq:likelihood-interaction}) was added for the tracked bubble volume. The developed algorithm performed well under the above conditions as well, as illustrated in Figures \ref{fig:tsl-split-merge}-\ref{fig:tsl-graph-entry-side} (the event in Figure \ref{fig:tsl-splerge} is a special case). As in the previous benchmark, split/merge events proximate in time were resolved (Figure \ref{fig:tsl-split-merge}) and the many-to-one and one-to-many events, much more frequent in this case, were also correctly identified (Figure \ref{fig:tsl-tri-merge}).

		\clearpage

		A very important special event is shown in Figure \ref{fig:tsl-splerge}, where the code detected a split event. However, it is, in fact, a very rapid, sub-resolution split event followed by a merge event. From the perspective of the available data, though, this is a many-to-many event that the present version of the code is not yet equipped to deal with. While in this benchmark only one such event occurs, many cases of bubble flow involve such types of interactions because the temporal resolution of experiments in general may not be sufficient to explicitly separate series of such events with high temporal density. This is especially challenging when some of the tracked properties are not sufficiently strongly conserved. We are currently developing a methodology that will enable MHT-X to treat such occurrences.

		\clearpage

		Figure \ref{fig:tsl-grapth-all} illustrates the trajectory graph for this benchmark. Note the more pronounced irregularity in trajectory patterns at the entry node (left) -- this has to do with very frequent merge events right after bubble takeoff at the TSL. The topside of the FOV (the right, exit node in Figure \ref{fig:tsl-grapth-all}) is dominated by split events, but these are much less frequent on average as their onset in the data series is delayed, while coalescence occurs regularly at the bottom of the FOV. Closeups of the entry node region are show in Figures \ref{fig:tsl-graph-entry} and \ref{fig:tsl-graph-entry-side} where bubble interactions can be seen in greater detail.

		In this case, although the bubble dynamics are somewhat more complicated for MHT-X to resolve despite the lower number density of objects compared to the first benchmark, the dataset is relatively clean in that the signal-to-noise ratio (SNR) in the original X-ray transmission images is rather high, and false positives are not present. To demonstrate the code's robustness against noisy input and false positives, we present the results from the third benchmark in the following section.

		\clearpage

		\subsection{Dynamic neutron radiography of bubble flow}

\begin{figure}[!h]
		        \centering			         
		        \includegraphics[width=1.0\linewidth]{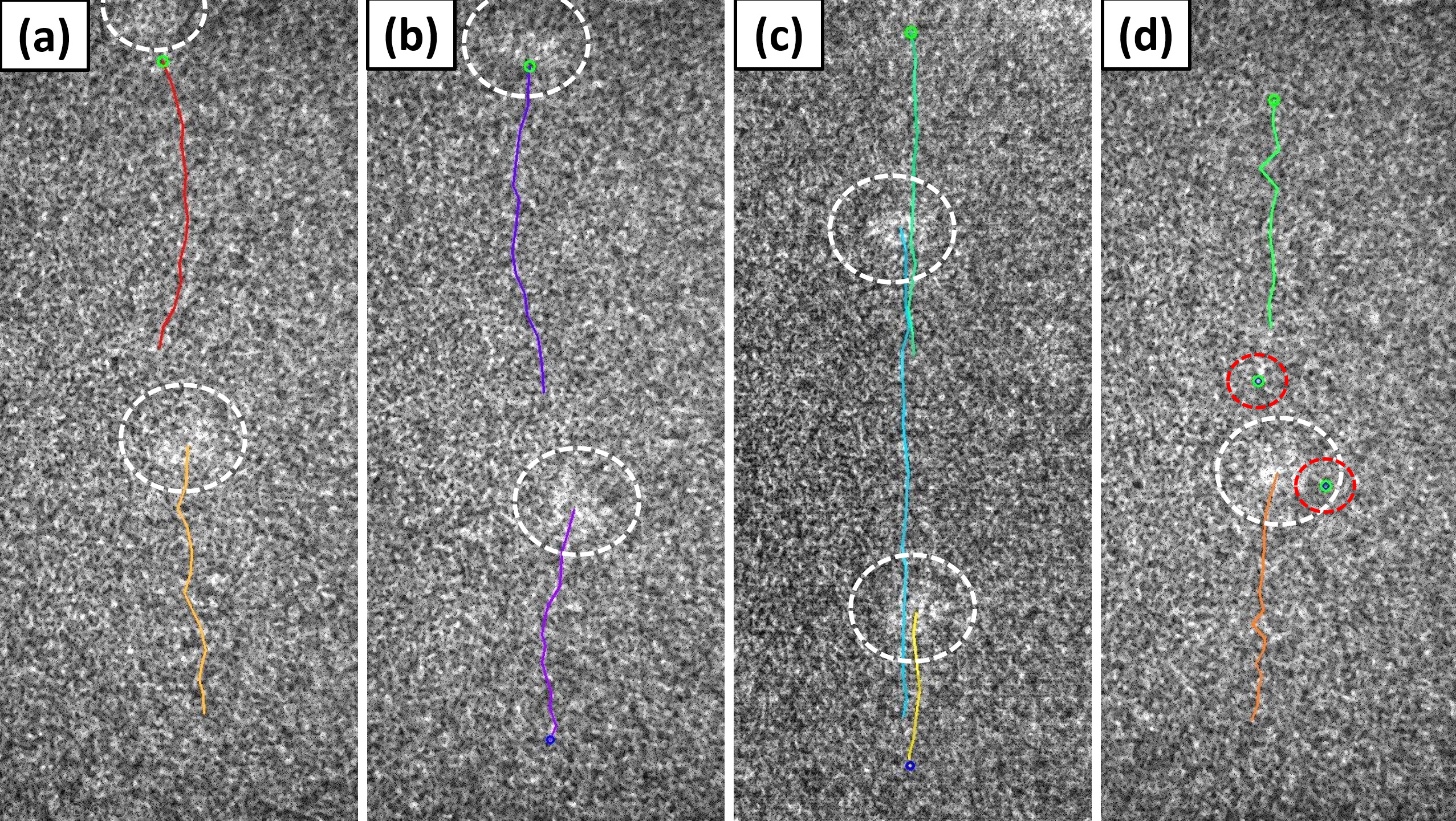}
    			\caption{Neutron radiography images with highlighted trajectories color coded by IDs, bubbles (white dashed circles) and false positives (red dashed circles). Up to a number of latest trajectory edges are shown in (a-d) for visual clarity.}
    			\label{fig:neutrons-images}
			\end{figure}

\begin{figure}[!h]
		        \centering			         
		        \includegraphics[width=1.0\linewidth]{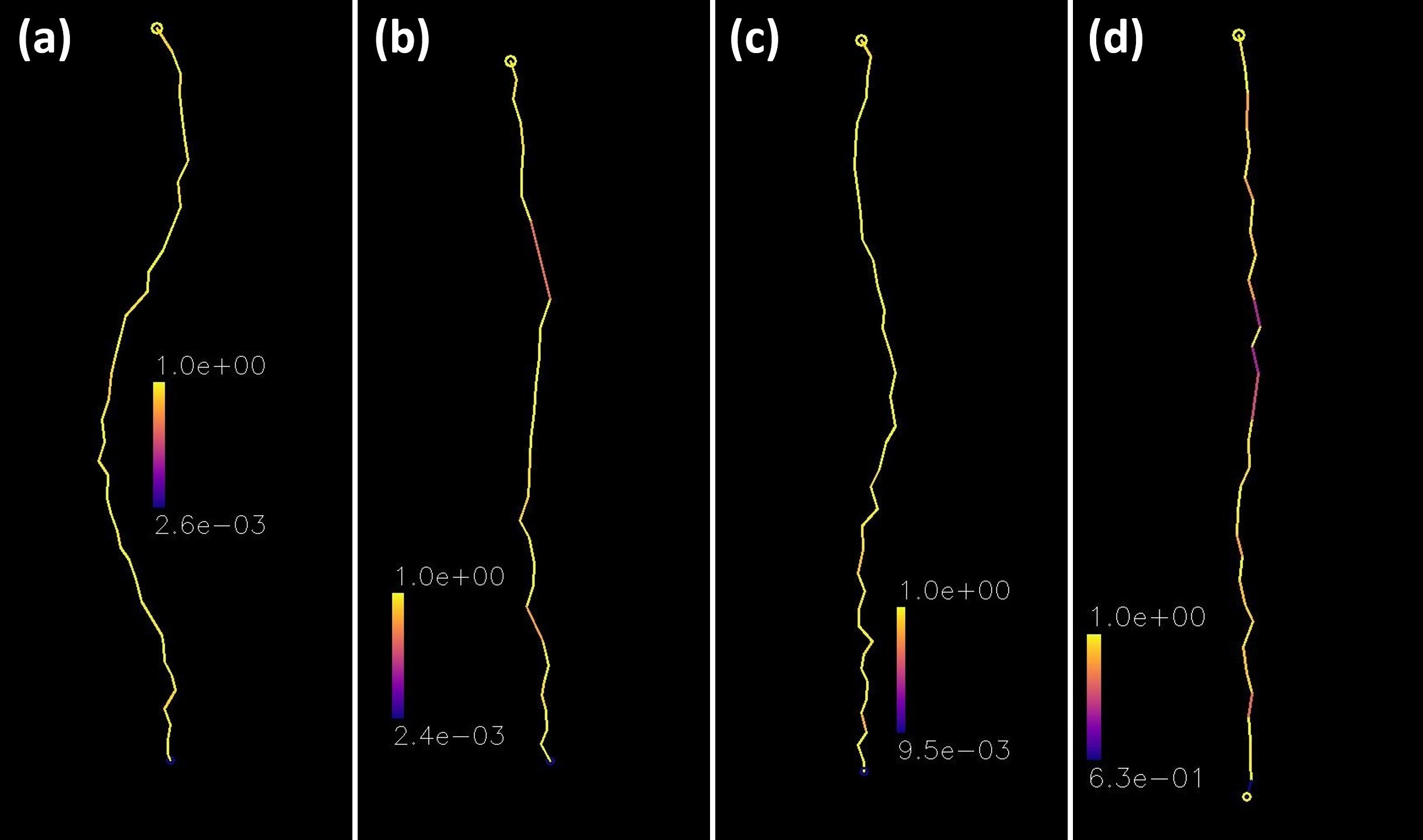}
    			\caption{Characteristic trajectories constructed by MHT-X for a sequence of neutron radiography images, with color coded trajectory edge probability.}
    			\label{fig:neutrons-trajectories}
			\end{figure}

		The third benchmark uses the data obtained by means of dynamic neutron imaging of argon bubble flow in liquid gallium in a system described in detail in \cite{birjukovsArgonBubbleFlow2020, birjukovsPhaseBoundaryDynamics2020}. The characteristic feature of neutron radiography images acquired with a high frame rate for thick liquid metal vessels is the very low image SNR. Because of this, even with advanced noise filtering and segmentation, the data provided as input to MHT-X is inevitably noisy in that bubble centroid position uncertainties are considerable and much greater than in the first two benchmarks shown herein. In addition, occasional false positives and detection failures may occur, further complicating tracing. To isolate these effects, image sequences with no bubble interactions were chosen, i.e. the average bubble spacing is sufficient to avoid collisions. Figures \ref{fig:neutrons-images} and \ref{fig:neutrons-trajectories} illustrate characteristic bubble trajectories reconstructed with MHT-X.

\begin{figure}[!h]
		        \centering			         
		        \includegraphics[width=0.65\linewidth]{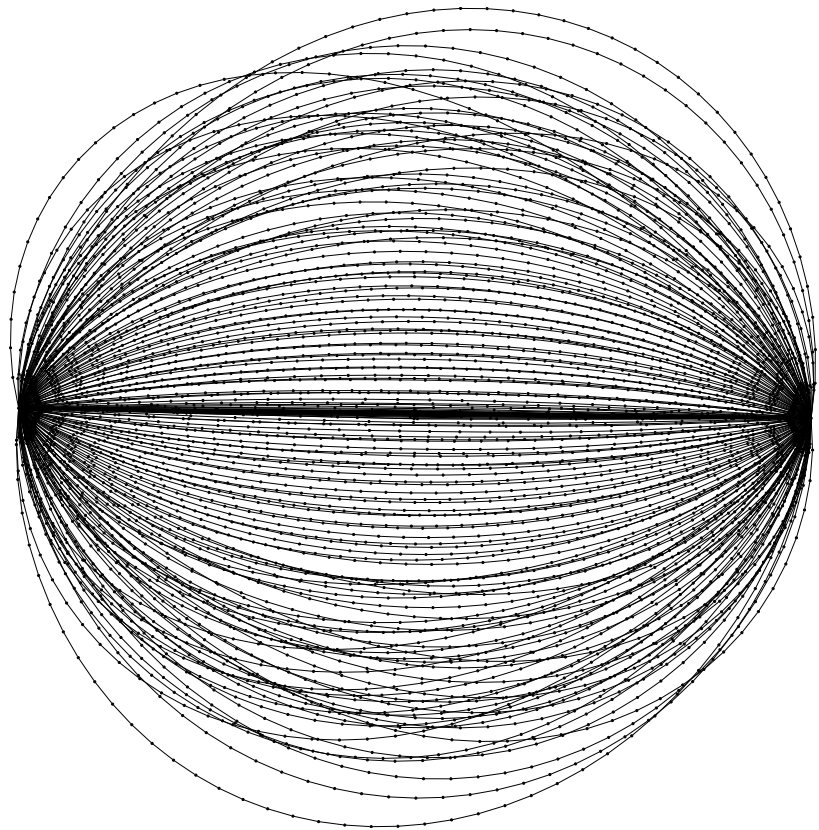}
    			\caption{A trajectory graph constructed for a sequence of neutron radiography images.}
    			\label{fig:neutrons-graph}
			\end{figure}

				One can clearly see that bubbles (white dashed circles in Figure \ref{fig:neutrons-images}) are largely shrouded by image noise. Despite this and the resulting noise in the object dataset, it is seen that the algorithm performs well and long, consistent trajectories spanning the entire FOV are recovered. Note also that in Figure \ref{fig:neutrons-images}d there are two false positives (red dashed circles) that were overlapping in time with true detections -- these were resolved as isolated nodes (1-node trajectories) in that within the solution graph they are only connected to the entry and exit nodes. In Figure \ref{fig:neutrons-trajectories}b, there is an edge with a lower probability in the upper part of the trajectory -- this is an example of MHT-X correctly extrapolating and connecting trajectory fragments across a frame with a detection failure event. Figure \ref{fig:neutrons-trajectories} also shows that the code is indeed resilient to noisy data, which is particularly evident in cases (c) and (d). The solution graph, in turn, is shown in Figure \ref{fig:neutrons-graph} where one can see how radically it differs from Figure \ref{fig:tsl-grapth-all} due to the absence of bubble interaction events. It must be noted that the graphs shown in this paper are not representative of data noise because of the way the graphs are transformed in \textit{GePhi}.

		\clearpage

		\section{Further extensions \& improvements}
		
		While the current version of the algorithm is already very versatile, as indicated by the above benchmark results, there are still potentially many ways it can be improved to make it more broadly applicable and enable solving tracing problems with greater particle density and more adverse conditions as seen in the cases outlined in the Introduction. The core components -- graph architecture, Bayesian MHT, Algorithm X -- must be backed by appropriate lower-level methods. To this end, it is planned to do the following in the future:
		
				\begin{itemize}
				    \item Replace the spline-based extrapolation with a Kalman filter predictor.
				    \item Develop a treatment for many-to-many object association events.
				    \item Implement a feedback loop that will enable coupling with image processing pipelines for iterative reinforced object detection and tracking.
				    \item Implement object cluster detection and joint tracking for improved tracing accuracy when dealing with swarms of coherently moving objects.
				    \item Introduce a scheme for edge probability re-evaluation (currently static once assigned) during successive graph sweeps to enable removing/reinforcing edges as more context becomes available after each time window sweep iteration.
				    \item Optimize performance: use a C-compiled graph computation package instead of the currently used \textit{Python} library; parallelize trajectory construction, association evaluation and visualization routines.
				\end{itemize}
		with potentially more modifications that currently are not considered. In addition, appropriate heuristics (statistical functions, etc.) for tracking of particles and other objects will be developed as necessary.

		\section{Conclusions \& outlook}

	We have developed and demonstrated the capabilities of an offline Bayesian multiple hypothesis tracker with a directed graph architecture that uses Algorithm X to solve the optimal association problem as an exact cover problem. The showcased benchmarks indicate that the current implementation is robust enough to process cases with relatively high object number density, in presence of data noise, false positives and detection failures. The algorithm is capable of resolving one-to-many split and many-to-one merge events for objects with variable shapes and parameters.
	
	In its current state and especially as the outlined improvements are implemented, we expect that the code will find use in the cases outlined in the Introduction and in other areas of research. Beyond that, the code is currently in use for the development of a dynamic mode decomposition code for the analysis of output of bubble flow simulations, as well as for bubble shape analysis, including shape evolution tracking and phase boundary velocimetry.
	
	The code is open source and is currently available on \textit{GitHub}: \url{https://github.com/Peteris-Zvejnieks/MHT-X}. It is frequently updated and a comprehensive documentation is also currently in the works.

    \section*{Acknowledgements}
    This research is a part of the ERDF project ”Development of numerical modelling approaches to study complex multiphysical interactions in electromagnetic liquid metal technologies” (No. 1.1.1.1/18/A/108). Neutron images were acquired at the Swiss spallation neutron source SINQ, Paul Scherrer Institute (PSI). The authors are grateful to Pavel Trtik (PSI), Jevgenijs Telicko (UL), Jan Hovind (PSI) and Knud Thomsen (PSI) for invaluable support in the neutron radiography experiments, and express gratitude to Natalia Shevchenko (HZDR) for her key role in the X-ray radiography experiments. The authors would like to acknowledge the \textit{math.stackexchange.com} user \textit{Watercrystal} for pointing out that the association optimization problem can be treated as an exact cover problem. Her involvement, while in the form of a single comment, was crucial to the development of this algorithm: \url{https://math.stackexchange.com/questions/3720630/finding-the-solution-set-for-binary-matrix-and-vector-multiplication/3720965#3720965}.

\clearpage

    \printbibliography[title={References}]

@article{birjukovsPhaseBoundaryDynamics2020,
	title = {Phase boundary dynamics of bubble flow in a thick liquid metal layer under an applied magnetic field},
	volume = {5},
	doi = {10.1103/PhysRevFluids.5.061601},
	journaltitle = {Physical Review Fluids},
	shortjournal = {Physical Review Fluids},
	author = {Birjukovs, Mihails and Dzelme, Valters and Jakovics, Andris and Thomsen, Knud and Trtik, Pavel},
	date = {2020-06-18}
}

@article{birjukovsArgonBubbleFlow2020,
	title = {Argon bubble flow in liquid gallium in external magnetic field},
	volume = {63},
	doi = {10.3233/JAE-209116},
	pages = {1--7},
	journaltitle = {International Journal of Applied Electromagnetics and Mechanics},
	shortjournal = {International Journal of Applied Electromagnetics and Mechanics},
	author = {Birjukovs, Mihails and Dzelme, Valters and Jakovics, Andris and Thomsen, Knud and Trtik, Pavel},
	date = {2020-04-29}
}

@article{megumi-x-rays,
author = {Akashi, Megumi and Keplinger, Olga and Shevchenko, Natalia and Anders, Sten and Reuter, Markus},
year = {2019},
month = {10},
pages = {},
title = {X-ray Radioscopic Visualization of Bubbly Flows Injected Through a Top Submerged Lance into a Liquid Metal},
volume = {51},
journal = {Metallurgical and Materials Transactions B},
doi = {10.1007/s11663-019-01720-y}
}

@article{megumi-cfd,
author = {Obiso, Daniele and Akashi, Megumi and Kriebitzsch, Sebastian and Meyer, B. and Reuter, Markus and Richter, Andreas},
year = {2020},
month = {06},
pages = {},
title = {CFD Modeling and Experimental Validation of Top-Submerged-Lance Gas Injection in Liquid Metal},
volume = {51},
journal = {Metallurgical and Materials Transactions B},
doi = {10.1007/s11663-020-01864-2}
}

@article{x-ray-bubble-breakup,
author = {Keplinger, Olga and Shevchenko, Natalia and Eckert, S.},
year = {2019},
month = {04},
pages = {39-50},
title = {Experimental investigation of bubble breakup in bubble chains rising in a liquid metal},
volume = {116},
journal = {International Journal of Multiphase Flow},
doi = {10.1016/j.ijmultiphaseflow.2019.03.027}
}

@article{x-ray-bubble-coalescence,
author = {Keplinger, Olga and Shevchenko, Natalia and Eckert, S.},
year = {2018},
month = {04},
pages = {159-169},
title = {Visualization of bubble coalescence in bubble chains rising in a liquid metal},
journal = {International Journal of Multiphase Flow},
volume = {105},
doi = {10.1016/j.ijmultiphaseflow.2018.04.001}
}

@article{neutrons-particles-lappan,
author = {Lappan, Tobias and Sarma, Martins and Heitkam, Sascha and Trtik, Pavel and Mannes, David and Eckert, Kerstin and Eckert, Sven},
year = {2020},
month = {10},
pages = {167-176},
title = {Neutron radiography of particle-laden liquid metal flow driven by an electromagnetic induction pump},
volume = {56},
journal = {Magnetohydrodynamics},
doi = {10.22364/mhd.56.2-3.8}
}

@article{neutrons-particles-stirrer-scepanskis,
author = {Sarma, Martins and Ščepanskis, Mihails and Jakovics, Andris and Thomsen, Knud and Nikoluškins, Raimonds and Vontobel, Peter and Beinerts, Toms and Bojarevics, Andris and Platacis, Erik},
year = {2015},
month = {09},
pages = {457-463},
title = {Neutron Radiography Visualization of Solid Particles in Stirring Liquid Metal},
volume = {69},
journal = {Physics Procedia},
doi = {10.1016/j.phpro.2015.07.064}
}

@article{neutrons-particles-stirrer-scepanskis-2,
author = {Ščepanskis, Mihails and Sarma, Martins and Vontobel, Peter and Trtik, Pavel and Thomsen, Knud and Jakovics, Andris and Beinerts, Toms},
year = {2017},
month = {04},
pages = {1045-1054},
title = {Assessment of Electromagnetic Stirrer Agitated Liquid Metal Flows by Dynamic Neutron Radiography},
volume = {48},
journal = {Metallurgical and Materials Transactions B},
doi = {10.1007/s11663-016-0902-8}
}

@article{neutrons-simulations-stirrer-valters,
author = {Dzelme, Valters and Jakovics, Andris and Vencels, Juris and Köppen, D. and Baake, E.},
year = {2018},
month = {10},
pages = {012047},
title = {Numerical and experimental study of liquid metal stirring by rotating permanent magnets},
volume = {424},
journal = {IOP Conference Series: Materials Science and Engineering},
doi = {10.1088/1757-899X/424/1/012047}
}

@article{neutrons-particles-froth-heitcam-tobias,
author = {Heitkam, Sascha and Lappan, Tobias and Trtik, Pavel and Eckert, Kerstin},
year = {2019},
month = {04},
pages = {},
title = {Tracking of Particles in Froth Using Neutron Imaging},
volume = {91},
journal = {Chemie Ingenieur Technik},
doi = {10.1002/cite.201800127}
}

@inproceedings{disjoint-graphs,
  author={G. {Brasó} and L. {Leal-Taixé}},
  booktitle={2020 IEEE/CVF Conference on Computer Vision and Pattern Recognition (CVPR)}, 
  title={Learning a Neural Solver for Multiple Object Tracking}, 
  year={2020},
  volume={},
  number={},
  pages={6246-6256},
  doi={10.1109/CVPR42600.2020.00628}}

@article{mht-deanonimized-main,
author = {Rubio, J.C. and Serrat, J. and López, Antonio},
year = {2012},
month = {01},
pages = {15-24},
title = {Multiple target tracking and identity linking under split, merge and occlusion of targets and observations},
volume = {2},
journal = {ICPRAM 2012 - Proceedings of the 1st International Conference on Pattern Recognition Applications and Methods},
doi = {10.5220/0003710600150024}
}

@inproceedings{mht-revisited,
author = {Kim, Chanho and Li, Fuxin and Ciptadi, Arridhana and Rehg, James},
year = {2015},
month = {12},
pages = {4696-4704},
title = {Multiple Hypothesis Tracking Revisited},
doi = {10.1109/ICCV.2015.533}
}

@article{mht-cox-efficient-implementation,
author = {Cox, I.J. and Hingorani, S.L.},
year = {1996},
month = {03},
pages = {138 - 150},
title = {An Efficient Implementation of Reid's Multiple Hypothesis Tracking Algorithm and Its Evaluation for the Purpose of Visual Tracking},
volume = {18},
journal = {Pattern Analysis and Machine Intelligence, IEEE Transactions on},
doi = {10.1109/34.481539}
}

@article{mht-reid-og,
  author={D. {Reid}},
  journal={IEEE Transactions on Automatic Control}, 
  title={An algorithm for tracking multiple targets}, 
  year={1979},
  volume={24},
  number={6},
  pages={843-854},
  doi={10.1109/TAC.1979.1102177}}

@article{mht-blackman,
author = {Blackman, Sam},
year = {2004},
month = {02},
pages = {5 - 18},
title = {Multiple Hypothesis Tracking for Multiple Target Tracking},
volume = {19},
journal = {Aerospace and Electronic Systems Magazine, IEEE},
doi = {10.1109/MAES.2004.1263228}
}

@online{knuthDancingLinks2000b,
  title = {Dancing Links},
  author = {Knuth, Donald E.},
  date = {2000-11-14},
  url = {http://arxiv.org/abs/cs/0011047},
  urldate = {2020-12-06},
  archivePrefix = {arXiv},
  eprint = {cs/0011047},
  eprinttype = {arxiv}
}

@article{sten-anders-vel-temp-measure,
author = {Anders, Sten and Noto, Daisuke and Tasaka, Yuji and Eckert, S.},
year = {2020},
month = {04},
pages = {},
title = {Simultaneous optical measurement of temperature and velocity fields in solidifying liquids},
volume = {61},
journal = {Experiments in Fluids},
doi = {10.1007/s00348-020-2939-3}
}

@article{sten-anders-spectral-random-masking,
author = {Anders, Sten and Noto, Daisuke and Seilmayer, Martin and Eckert, S.},
year = {2019},
month = {03},
pages = {68},
title = {Spectral random masking: a novel dynamic masking technique for PIV in multiphase flows},
volume = {60},
journal = {Experiments in Fluids},
doi = {10.1007/s00348-019-2703-8}
}

\end{document}